\documentclass[conference]{IEEEtran}
\usepackage{microtype}
\usepackage{amsfonts}
\usepackage{graphicx,subfigure}
\usepackage{multirow}
\usepackage{url}
\usepackage[hang,flushmargin]{footmisc} 
\usepackage{color}
\usepackage{tablefootnote}

\begin{document}
\title{Classification of Radiology Reports\\Using Neural Attention Models}

\author{\IEEEauthorblockN{Bonggun Shin$^{1}$, Falgun H. Chokshi$^{2,}$$^{3}$, Timothy Lee$^{1}$, Jinho D. Choi$^{1}$}
\IEEEauthorblockA{$^{1}$Mathematics and Computer Science, Emory University, Atlanta, GA 30322 \\
$^{2}$Radiology and Imaging Sciences, Emory University School of Medicine, Atlanta, GA 30322\\
 $^{3}$Biomedical Informatics, Emory University School of Medicine, Atlanta, GA 30322\\
Email: bonggun.shin@emory.edu, fchoksh@emory.edu, timlee1028@gmail.com, jinho.choi@emory.edu}
}


%


\maketitle

\begin{abstract}
The electronic health record (EHR) contains a large amount of multi-dimensional and unstructured clinical data of significant operational and research value. Distinguished from previous studies, our approach embraces a double-annotated dataset and strays away from obscure ``black-box'' models to comprehensive deep learning models. In this paper, we present a novel neural attention mechanism that not only classifies clinically important findings. Specifically, convolutional neural networks (CNN) with attention analysis are used to classify radiology head computed tomography reports based on five categories that radiologists would account for in assessing acute and communicable findings in daily practice. The experiments show that our CNN attention models outperform non-neural models, especially when trained on a larger dataset. Our attention analysis demonstrates the intuition behind the classifier's decision by generating a heatmap that highlights attended terms used by the CNN model; this is valuable when potential downstream medical decisions are to be performed by human experts or the classifier information is to be used in cohort construction such as for epidemiological studies. 
\end{abstract}


%
\IEEEpeerreviewmaketitle

\section{Introduction}
\label{sec:intro}

\noindent Electronic health systems (EHR) are replete with large volumes of unstructured data that can be mined for useful population and patient level information~\cite{mcafee2012big}. With increased mandates by federal regulators to demonstrate quality, improve outcomes, and reduce costs~\cite{burwell2015setting}, there is an increasing need to develop scalable and reliable methods of unstructured data mining. Additionally, the Precision Medicine Initiative (PMI)~\cite{collins2015new} has spearheaded the need for powerful text mining techniques to promote more nuanced phenotyping of patients and patient populations~\cite{simmons2016text}. 

EHR data is comprised of both structured (e.g. lab values, vital signs) and unstructured (e.g. clinical notes, radiology reports) text elements. This unstructured data contains rich information that could be used for many purposes if automated text analysis systems were developed. 
Recent studies have attempted to derive structures from such unstructured clinical notes to 
evaluate cancer treatment outcomes~\cite{mathias2012use}, 
identify patient phenotype cohorts~\cite{shivade2014review,zhou2014mining},
or predict clinical outcomes~\cite{staff2013can}.
Most of these methods either devise query-based approaches or develop rules-based approaches, which are often impractical because these approaches do not consider contextual information about the keywords presented in the texts.
For example, on a radiology report of a head computed tomography (CT) scan, an attempt to categorize bleeding patients with the query word ``hemorrhage'' will fail because the result could also contain false negative cases where the usage of the query word is in the opposite context such as ``no more hemorrhage''.

To ameliorate this problem,
more sophisticated approaches using natural language processing (NLP) such as
an $n$-gram model~\cite{marafino2014n} or 
a pipeline of NLP components~\cite{savova2010discovering,afzal2016identifying} have been proposed.
Although these methods
shed light on extracting partial information from clinical notes,
three drawbacks should be addressed 
in order to create a more practical model:

\begin{itemize}
\item Capturing multifaceted information still falls short compared to human performance~\cite{perera2013challenges}.
\item Although deep learning significantly outperforms other conventional methods in many domains such as computer vision~\cite{krizhevsky2012imagenet,sermanet2013pedestrian}, speech recognition~\cite{dahl2012context,graves2013speech}, sentiment analysis~\cite{shin:16a,poria2015deep}, etc.,
previous studies on classification of clinical notes have relied on datasets that are too small for deep learning techniques to be effective.
For instance, Savova et al.~\cite{savova2010discovering} experimented with only 550 clinical notes in their research.
\item As Girshick et al.~\cite{girshick2015deformable} noted, neural network models are ``black-box'' methods because it is nearly  impossible to know how the machine produces a specific output. For its lack of interpretability, human cannot judge if the output of the model is trustworthy.
\end{itemize}

\noindent To overcome these issues, we first construct a Convolutional Neural Network (CNN) model specifically designed for document classification, that is similar to the one employed by Kim \cite{kim2014convolutional}.
Unlike traditional bag-of-words approaches taking $n$-grams in a sparse vector format, this CNN model takes input text in a dense vector format using word embeddings~\cite{mikolov2013efficient}.
We then introduce an efficient attention mechanism to our CNN model that provides a global view of the document by emphasizing (or de-emphasizing) important words (Section~\ref{ssec:embedding-attention}).

Our models are evaluated on radiology head CT reports from intensive care unit (ICU) patients with altered mental status,
which are annotated by two experienced practicing attending physicians in radiology and adjudicated by a radiologist. We focus on radiology reports because they offer a major source of unstructured data that could be mined and applied towards predictive models, which could assess outcomes such as length of stay, mortality, resource utilization, and cost-analysis. 
The annotated dataset created for this project is large enough for deep learning techniques to be effective.
Our experiments show that the CNN model outperforms other machine learning models using linear classification and random forest  (Section~\ref{sec:experiments}). Moreover, our research further adds interpretability to the data by applying an attention mechanism to the CNN model.
To the best of our knowledge, this is the first time that an attention mechanism is introduced for classifying radiology reports.\footnote{All our resources will be publicly available upon acceptance.}

\section{Related Work}
\label{sec:related-work}

\noindent Methods of extracting unstructured information from the EHR traditionally focused on rule-based systems of NLP, machine learning and statistical analysis, a hybrid of these systems, or cohort identification systems~\cite{shivade2014review}. Regarding Machine Learning and Statistical analysis, Kawaler et al.~\cite{kawaler2012} reports promising results on predicting post-hospitalization venous thromboembolism (VTE) risk from EHRs by using general machine learning techniques such as Naive Bayes, Support Vector Machines (SVM), \textit{k}-nearest neighbor (\textit{k}-NN), and Random Forest. Marafino et al.~\cite{marafino2014n} also successfully used $n$-gram SVM to help clinical diagnosis classification in ICU. Applications of neural networks also gained tremendous momentum in clinical note extraction, especially in relation extraction and named entity recognition. CNN, although originally invented for the purpose of solving computer vision, has proven to work profoundly well in various NLP tasks, and used for supervised learning and automatically learning features for classification of relation extraction~\cite{liu2013convolution} and named entity recognition~\cite{lample2016neural}.

CNN has also seen upsurge in popularity in document level text classification such as sentiment analysis and question answering
~\cite{kim2014convolutional,Severyn2015TwitterSA,jurczyk:16a}. A more recent approach in clinical and biomedical document classification relies on a CNN model proposed by Kim \cite{kim2014convolutional}, and leverages the CNN's convolution feature and its ability to effectively capture both semantic and syntactic information to gain a solid 3\% boost in F1 score over prior results \cite{Rios2015ConvolutionalNN}.

The attention mechanism is a method of emphasizing or de-emphasizing features that are more or less important in neural network classification problems \cite{attention2014}. Originally developed for image processing, attention mechanism has successfully been adopted in various NLP domains including question answering, sentiment analysis, machine translation, and document level classification ~\cite{Shih2015WhereTL,Stollenga2014DeepNW,Cho2014LearningPR,shin:16a}. The attention mechanism introduced here is efficient and gives a comprehensive way of understanding the classification decision.





\section{Approach}
\label{sec:approach}

\noindent This section first describes baseline methods using several bag-of-words representations (BOW) coupled with linear classifiers such as logistic regression and support vector machines (SVM), and a non-linear classifier such as random forest (Section~\ref{ssec:baseline}).
We then depict a Convolutional Neural Network (CNN) model using word embeddings from different distributional semantics methods (Section~\ref{ssec:cnnmodel}).
Finally, we elaborate how our attention mechanism is incorporated into the CNN model (Section~\ref{ssec:embedding-attention}).


\subsection{Baseline Methods}
\label{ssec:baseline}

\noindent To establish strong baselines, non-neural classifiers using BOW are experimented, which give competitive performance to other complex models although their model complexities are lower.
These baseline models are selected to contrast the performance of the proposed CNN models in Sections~\ref{ssec:cnnmodel} and \ref{ssec:embedding-attention}.
\vspace{1ex}

\noindent\textit{Vector Representations}\vspace{1ex}

\noindent Four types of vector space models are used to represent BOW, where each term $w_i$ in a document $d_j \in D$ is represented by:

\begin{enumerate}
\item Term frequency\\: $t_{ij}$ = \# of times that $w_i$ occurs in $d_j$
\item Term frequency normalized by the document size\\: $\frac{t_{ij}}{\sum_{\forall w_k \in d_j} t_{kj}}$
\item Binary representation of the term\\: $t_{ij} = 1$ if $w_i$ occurs at least once in $d_j$; otherwise, $0$
\item Term frequency inverse document frequency (TF-IDF)\\:$t_{ij} \cdot \log\frac{|D|}{|d\in D~:~w_i \in d|}$
\end{enumerate}\vspace{1ex}

\noindent Stopwords are removed for the first three models, whereas they are not removed for the last model because TF-IDF implicitly filters those out by assigning lower weights.\footnote{We used the stopword list provided by the open source NLP toolkit, NLP4J\\: \texttt{http://github.com/emorynlp/nlp4j}}
\vspace{1ex}

\noindent\textit{Non-neural Classifiers}\vspace{1ex}

\noindent Various non-neural classifiers such as SVM using the hinge loss, logistic regression using the log loss, and random forest are used to build the baseline models.
For experiments, implementations of these classifiers in \texttt{scikit-learn} are used.\footnote{\texttt{scikit-learn}: \texttt{http://scikit-learn.org}} 


\subsection{Convolutional Neural Networks}
\label{ssec:cnnmodel}

\begin{figure}[!htbp]
\centering
\includegraphics[width=\linewidth]{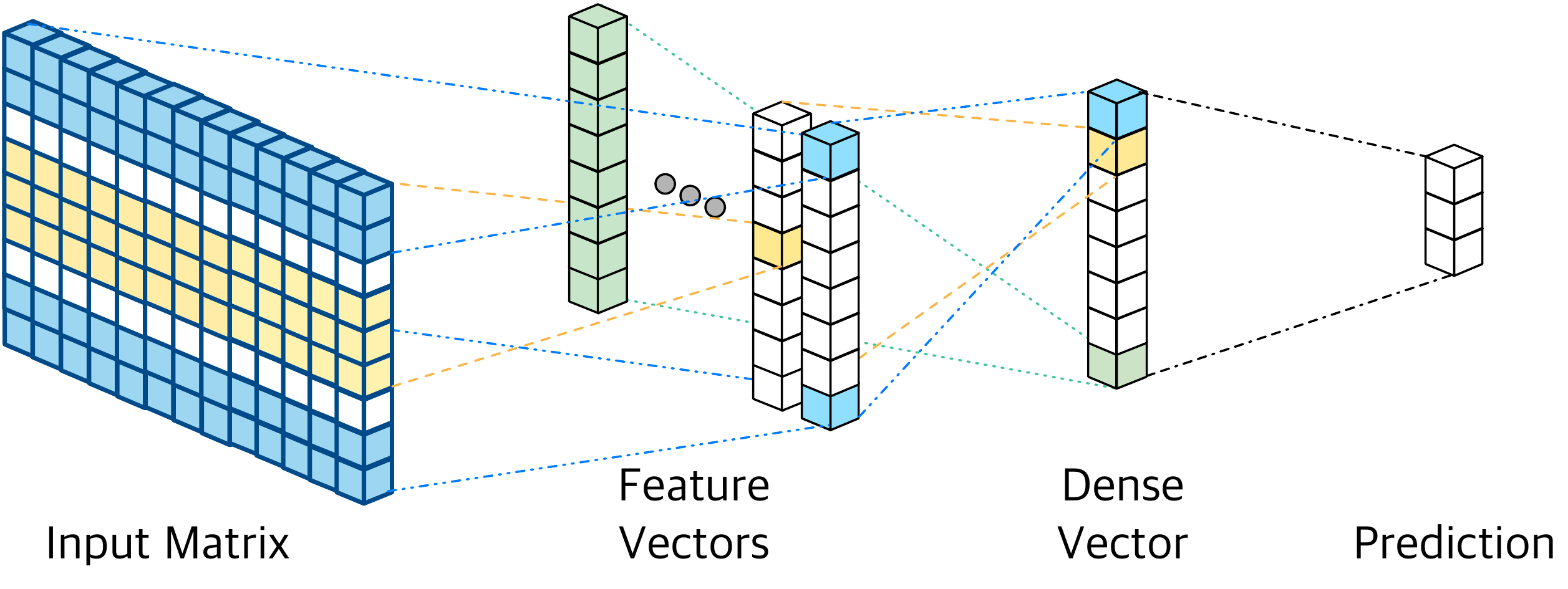}
\caption{The overview of our CNN model for document classification.}
\label{fig:cnn}
\end{figure}

\noindent Our first approach is a single-layer CNN model (Figure~\ref{fig:cnn})
using pre-trained word embeddings, 
which is a mirror implementation of the CNN model 
introduced by Kim \cite{kim2014convolutional}. 
Let $s \in \mathbb{R}^{n \times d}$ be a matrix representing the input document, where $n$ is the number of words, $d$ is the dimension of the word embeddings, and each row corresponds to the word embedding, $w_i \in \mathbb{R}^d$, where $w_i$ indicates the $i$'th word in the document.
A word embedding can be learned by either continuous bag-of-words (CBOW) or skip-gram (SKIP) models.
While CBOW learns a proper word vector for a given set of words in context, SKIP is trained to predict a vector representing neighboring words for an input word. 
Since each embedding model has its own strength~\cite{mikolov2013distributed}, both models are considered for the best configuration.

The document matrix $s$ made of any of these embeddings is fed into the convolutional layer and convolved by the weights $c\in \mathbb{R}^{l \times d}$, where $l$ is the length of the filter.
The convolutional layer can take $m$-number of filters of the length $l$.
Each convolution produces a vector $v_c \in \mathbb{R}^{n-l+1}$, where elements in $v_c$ convey the $l$-gram features across the document.
The max pooling layer selects the most salient features from each of the $m$ vectors produced by the filters. 
As a result, the output of this max pooling layer is a vector $v_m \in \mathbb{R}^{(n-l+1) \times m}$. 
The selected features are passed onto the softmax layer, which is optimized for the score of each sentiment class label.

\subsection{Embedding Attention}
\label{ssec:embedding-attention}

\begin{figure}[ht!]
\centering     
\subfigure[Given a document matrix, the attention matrix is first created by performing multiple convolutions. The attention vector is then created by performing max pooling on each row of the attention matrix.\vspace{2ex}]
{\label{fig:eav1}\includegraphics[scale=0.3]{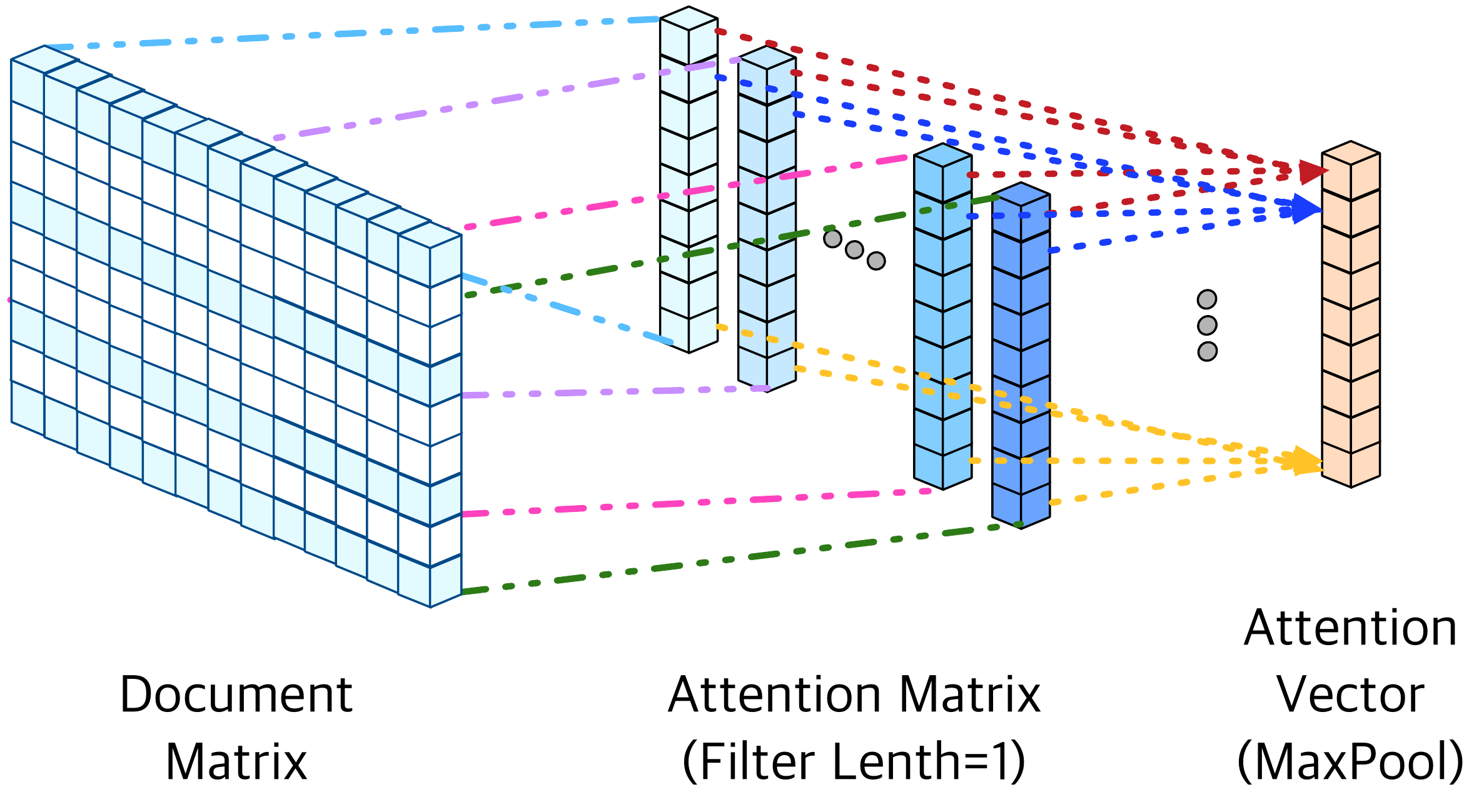}}
\subfigure[The embedding attention vector is created by multiplying the transposed document matrix to the attention vector.]
{\label{fig:eav2}\includegraphics[scale=0.3]{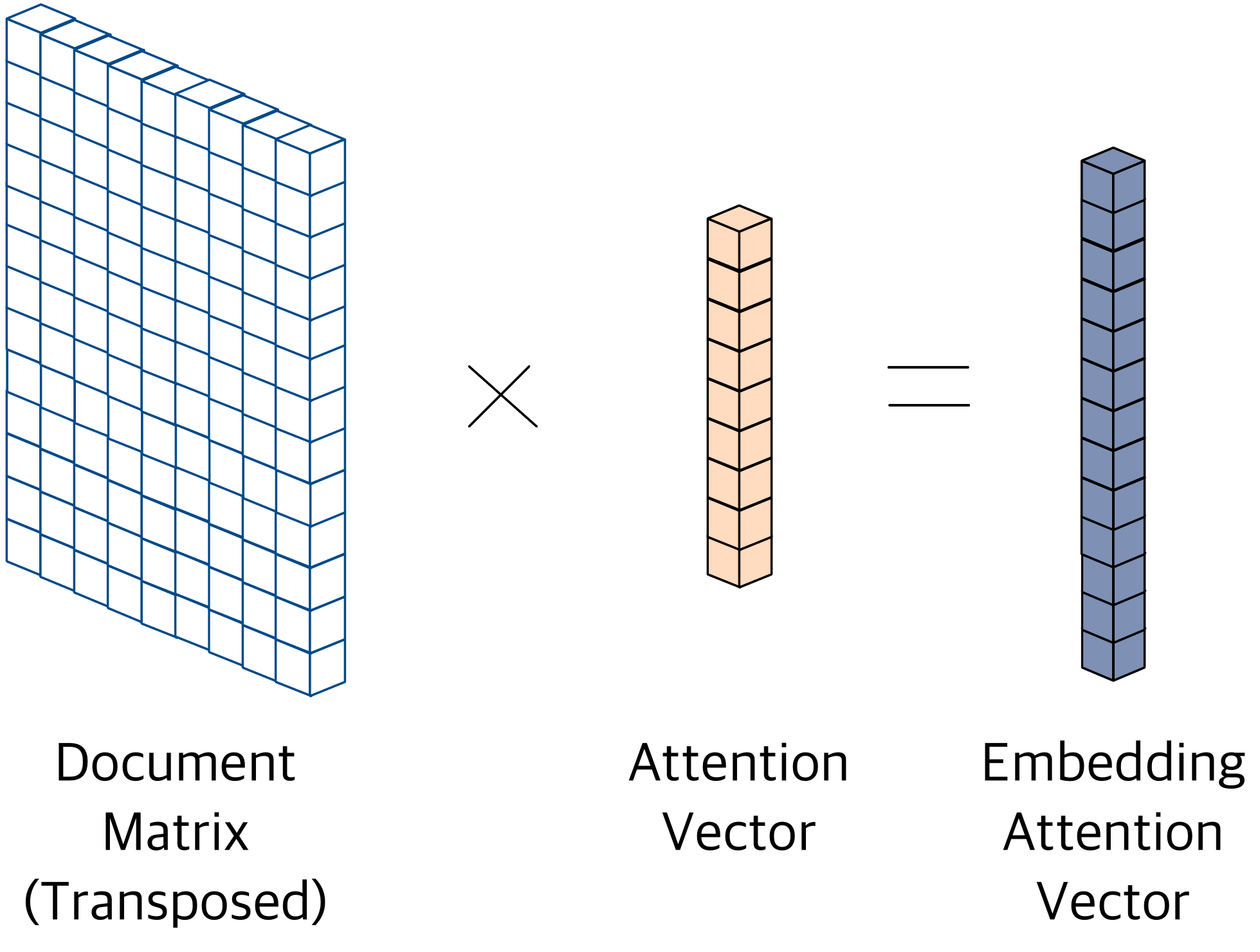}}
\caption{Construction of the embedding attention vector from a doc.\ matrix.}
\label{fig:embedding-attention}
\end{figure}

\noindent The CNN model uses several filters with different lengths; given the filter length $l$, the convolution considers $l$-gram features.
However, these $l$-gram features account only for local views, not the global view of the document, which is necessary for several transitional cases such as negation in sentiment analysis~\cite{socher2012semantic}.
To ameliorate this issue, we introduce the embedding attention vector (EAV), which transforms the document matrix into a vector.
For example, the EAV is calculated as a weighted sum of each column in the document matrix $s \in \mathbb{R}^{n \times d}$, which yields a vector $v \in \mathbb{R}^{d}$.
For each document, one EAV can be derived from the document matrix that contains attention information.
The document matrix are used to create the EAV through multiple convolutions and max pooling as follows:

\vspace{1ex}
\begin{enumerate}
\item Apply $m$-number of convolutions with the filter length $1$ to the document matrix $s \in \mathbb{R}^{n \times d}$. 
\item Aggregate all convolution outputs to form an attention matrix $s_a \in \mathbb{R}^{n \times m}$, where $n$ is the number of words in the document, and $m$ is the number of filters whose length is $1$.
\item Execute max pooling for each row of the attention matrix $s_a$, which generates the attention vector $v_a \in \mathbb{R}^n$ (Figure~\ref{fig:eav1}).
\item Transpose the document matrix $s$ such that $s^T \in \mathbb{R}^{d \times n}$, and multiply it with the attention vector $v_a \in \mathbb{R}^{n}$, which generates the embedding attention vector $v_e \in \mathbb{R}^d$ (Figure~\ref{fig:eav2}).
\end{enumerate}
\vspace{1ex}

\noindent The resulting EAVs are appended to the penultimate layer to serve as additional information for the softmax layer.
It is worthy to note that 
the proposed model is an additive model,  
where the network can be seen as a two-pathways network.
Although this simplification is desirable 
in terms of speed, 
multiplicative attentions might be more appropriate 
if focusing on the performance.

\section{Experiments}
\label{sec:experiments}


\subsection{Corpus}
\label{ssec:corpora} 

\noindent All models are experimented on radiology head CT reports of patients from intensive care units (ICUs) with altered mental status.
The dataset is provided by Emory Healthcare after Institutional Review Board approval; given the dataset, we create a new corpus where each report is annotated by two experienced practicing attending physicians and adjudicated by a radiologist such that the inter-annotator agreement (ITA) can be measured.
Each report is manually annotated for five classification tasks, where each task involves three labels implying the degree of the severity, as adapted from~\cite{chokshi2016altered}.
These five tasks are as follows:\footnote{The authors plan to make the de-identified version of this corpus available.}

\vspace{1ex}
\begin{enumerate}
\item \textit{Severity of Study} - 0: normal, 1: abnormal study, but no acute or communicable findings, 2: abnormal Study, with acute and communicable findings.
\item \textit{Acute Intracranial Bleed} - 0: not present, 1: present, but not new or worse, 2: new or worse.
\item \textit{Acute Mass Effect} (herniation) - 0: not present, 1: present, but not new or worse, 2: new or worse.
\item \textit{Acute Stroke} - 0: not present, 1: present, but not new or worse, 2: new or worse.
\item \textit{Acute Hydrocephalus} (ventriculomegaly) - 0: not present, 1: present, but not new or worse, 2: new or worse.
\end{enumerate}\vspace{1ex}


\begin{table*}[ht]
\centering
\caption{Accuracy (in \%) of the baseline models using different combinations of classifiers and vector representations on the five tasks. }
\label{tbl:baseline}
\begin{tabular}{c||c|c|c|c||c|c|c|c||c|c|c|c}
        & \multicolumn{4}{c||}{\textbf{Logistic Regression}}      & \multicolumn{4}{c||}{\textbf{Random Forest}} & \multicolumn{4}{c}{\textbf{Support Vector Machines}}     \\
        & TF   & TF-Norm       & Binary        & TF-IDF & TF    & TF-Norm  & Binary & TF-IDF & TF            & TF-Norm & Binary & TF-IDF        \\ \hline \hline
Task 1  & 83.0 & 83.0          & 80.0          & 83.0   & 81.0  & 77.5     & 78.0   & 81.0   & 81.0          & 83.0    & 78.0   & \textbf{85.5} \\
Task 2  & 79.0 & 82.5          & 77.0          & 79.5   & 76.0  & 75.5     & 74.0   & 76.0   & 81.5          & 82.5    & 73.5   & \textbf{83.0} \\
Task 3  & 82.0 & 81.5          & 75.5          & 81.0   & 76.5  & 76.5     & 74.0   & 74.5   & 80.5          & 81.0    & 76.0   & \textbf{83.5} \\
Task 4  & 87.0 & \textbf{87.5} & \textbf{87.5} & 87.0   & 81.5  & 80.0     & 81.5   & 81.0   & 85.0          & 86.0    & 84.0   & 85.5          \\
Task 5  & 80.5 & 82.5          & \textbf{83.5} & 83.0   & 75.0  & 75.5     & 74.5   & 75.0   & \textbf{83.5} & 81.5    & 80.0   & 81.0          \\ \hline \hline
Average & 82.3 & 83.4          & 80.7          & 82.7   & 78.0  & 77.0     & 76.4   & 77.5   & 82.3          & 82.8    & 78.3   & \textbf{83.7}
\end{tabular}
\end{table*}

\begin{figure}[!htbp]
\centering
\includegraphics[width=0.8\linewidth]{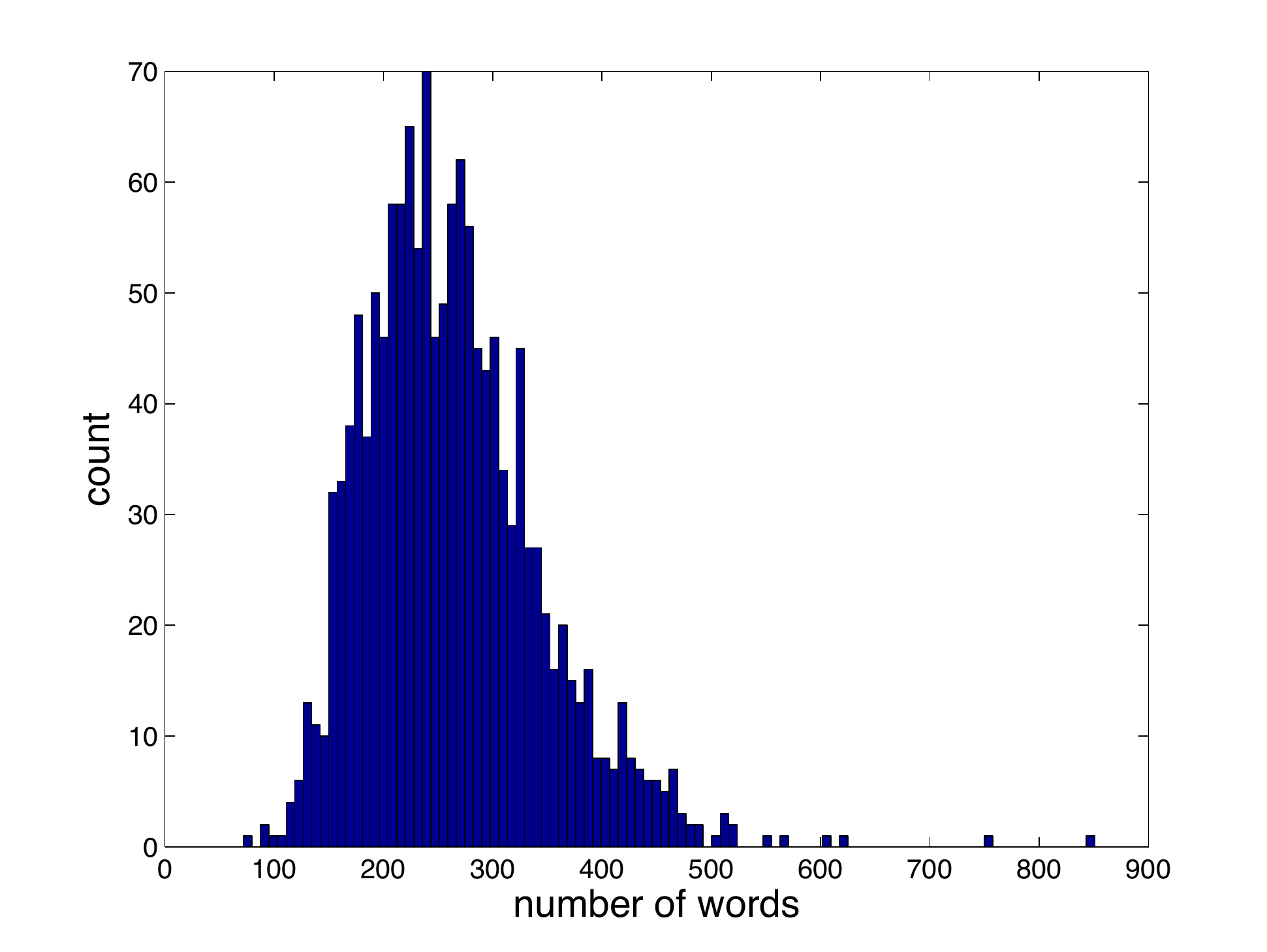}
\caption{The histogram of the word counts for each radiology report, which ranges between 72 and 851.}
\label{fig:hist}
\end{figure}

\noindent Table~\ref{tbl:raddata} shows the statistics of the radiology head CT reports for each classification task.
For each task, the dataset is split into training, development, and evaluation sets (1000/200/200), where each label is proportionally distributed in each set.
As noted in Section~\ref{ssec:cnnmodel},
the number of words in each document, $n$,
needs to be fixed 
such that the output of each convolution layer
stays the same.
After examining the histogram that shows
the distribution of 
the word counts for each radiology report 
(Figure~\ref{fig:hist}),
$n=800$ is picked. 
Although the word count ranges between 72 and 851,
extreme outliers are excluded when choosing $n$.


\begin{table}[htbp!]
\centering
\caption{Statistics of the radiology head CT reports for each task. Each column shows the number of reports in each category with respect to the degree of the severity.}
\label{tbl:raddata}
\begin{tabular}{l||r|r|r||r}
 & \multicolumn{1}{c|}{\bf 0} & \multicolumn{1}{c|}{\bf 1} & \multicolumn{1}{c||}{\bf 2} & \multicolumn{1}{c}{\bf All} \\
\hline\hline
\texttt{Severity of Study} & 58 &  940 & 402 & 1,400 \\
\texttt{Acute Blood} &   653 &    546 &   201 &  1,400 \\
\texttt{Mass Effect} & 751 & 443 & 206 & 1,400 \\ 
\texttt{Acute Stroke} & 1,113 & 173 & 114 & 1,400 \\ 
\texttt{Hydrocephalus} & 1,078 & 172 & 150 & 1,400 \\
\end{tabular}
\end{table}


\subsection{Word Embedding Construction}

\noindent To best capture the word semantics in the radiology domain, 80,000 head CT reports without manual annotation are used to train word embeddings.
We vary the number of radiology reports during training so that the impact of bigger unstructured training data for building word embeddings can be analyzed for the task of document classification in radiology reports (see details in Section~\ref{sssec:cnnmodel}).
All documents are pre-tokenized by the open-source toolkit, NLP4J.
The word embeddings are trained by the original implementation of \texttt{word2vec}~\cite{mikolov2013distributed,mikolov2013efficient} using CBOW and SKIP models and negative sampling.\footnote{\texttt{word2vec}: \texttt{http://code.google.com/p/word2vec}}
No explicit hyper-parameter tuning is performed.
Three sets of embeddings with different dimensions (100, 200, 400) are trained to observe the impact of the embedding size on each approach. 


\subsection{Evaluation}
\label{ssec:evaluation}

\noindent To demonstrate the superiority of the proposed neural methods, the performance results from the baseline models in Section~\ref{ssec:baseline} are first presented (Section~\ref{sssec:baseline}).
For the CNN model proposed in Section~\ref{ssec:cnnmodel}, the best hyper-parameter configuration is found through grid search on each development set.
Although our grid search is not exhaustive, meaningful trends of performances are found and reported in Section~\ref{sssec:cnnmodel}.
The attention enabled CNN model
successfully presented
rationales for the corresponding decisions.
We visualize 
this machine generated explanation 
as a heatmap overlayed on the report
in Section~\ref{sssec:neural-attention-model}.

We analyze 
the results between two proposed CNN models and the baseline models
and show the effectiveness of deep learning 
on document classification of radiology reports
and the practicality of the interpretable neural model.
These models include
logistic regressions, SVM, and random forest (baseline, Section~\ref{ssec:baseline}), 
plane CNN (CNN; Section~\ref{ssec:cnnmodel}), 
and 
CNN with the neural attention mechanism (NAM; Section~\ref{ssec:embedding-attention}).
The model selection of all neural models is carried with
three types of data split:
training, development, and evaluation sets.
After different models learn from training data,
the best model is selected based on
the performance tested on the development set, then the final score is reported using the evaluation set.

\begin{table*}[!ht]
\centering
\caption{Accuracy (in \%) of our CNN models using different sets of hyper-parameters measured on the development set. 
The best model for each task is marked in bold text.
Models vary in configurations of different word2vec settings,
such as the dimension of word embeddings (W2V-DIM), the number of documents used for embedding training
(W2V-ND), 
and the optimization methods for embedding training (SKIP and CBOW).
}
\label{tbl:cnn}
\begin{tabular}{c||c|c|c|c|c|c|c|c|c|c|c|c}
W2V-TASK & \multicolumn{12}{c}{SKIP}                                                                                   \\ \hline \hline
W2V-DIM  & \multicolumn{4}{c|}{100}  & \multicolumn{4}{c|}{200}           & \multicolumn{4}{c}{400}                    \\ \hline
W2V-ND   & 20k  & 40k  & 60k  & 80k  & 20k  & 40k  & 60k  & 80k           & 20k  & 40k  & 60k           & 80k           \\ \hline \hline
Task 1   & 84.0 & 87.5 & 88.5 & 88.0 & 84.5 & 86.5 & 88.0 & \textbf{89.0} & 84.0 & 87.5 & 88.5          & 88.0          \\
Task 2   & 82.0 & 87.5 & 88.5 & 87.5 & 82.5 & 87.0 & 89.0 & 88.0          & 83.0 & 88.0 & 89.0          & \textbf{90.0} \\
Task 3   & 82.5 & 86.5 & 85.0 & 85.0 & 84.5 & 86.0 & 87.0 & 86.5          & 82.5 & 87.0 & \textbf{87.5} & 86.5          \\
Task 4   & 89.0 & 92.0 & 92.0 & 92.0 & 87.5 & 92.0 & 91.5 & 92.0          & 87.0 & 92.0 & 91.5          & 91.5          \\
Task 5   & 88.0 & 90.5 & 92.0 & 92.0 & 90.0 & 91.0 & 91.5 & \textbf{92.5} & 87.0 & 91.0 & 91.5          & 91.0         
\end{tabular}
\vspace{2ex}
\\
\begin{tabular}{c||c|c|c|c|c|c|c|c|c|c|c|c}
W2V-TASK & \multicolumn{12}{c}{CBOW}                                                                                                                                                                                                                                                                                                                           \\ \hline \hline
W2V-DIM  & \multicolumn{4}{c|}{100}                                                                                      & \multicolumn{4}{c|}{200}                                                                                                        & \multicolumn{4}{c}{400}                                                                          \\ \hline
W2V-ND   & \multicolumn{1}{c|}{20k}  & \multicolumn{1}{c|}{40k}  & \multicolumn{1}{c|}{60k}  & \multicolumn{1}{c|}{80k}  & \multicolumn{1}{c|}{20k}  & \multicolumn{1}{c|}{40k}           & \multicolumn{1}{c|}{60k}  & \multicolumn{1}{c|}{80k}           & \multicolumn{1}{c|}{20k}  & \multicolumn{1}{c|}{40k}  & \multicolumn{1}{c|}{60k}  & 80k           \\ \hline \hline
Task 1   & \multicolumn{1}{c|}{84.0} & \multicolumn{1}{c|}{88.5} & \multicolumn{1}{c|}{88.5} & \multicolumn{1}{c|}{88.5} & \multicolumn{1}{c|}{82.0} & \multicolumn{1}{c|}{87.0}          & \multicolumn{1}{c|}{87.5} & \multicolumn{1}{c|}{87.5}          & \multicolumn{1}{c|}{84.0} & \multicolumn{1}{c|}{86.5} & \multicolumn{1}{c|}{87.5} & \textbf{89.0} \\
Task 2   & \multicolumn{1}{c|}{82.0} & \multicolumn{1}{c|}{87.0} & \multicolumn{1}{c|}{88.5} & \multicolumn{1}{c|}{88.5} & \multicolumn{1}{c|}{84.0} & \multicolumn{1}{c|}{87.0}          & \multicolumn{1}{c|}{89.5} & \multicolumn{1}{c|}{88.0}          & \multicolumn{1}{c|}{84.0} & \multicolumn{1}{c|}{86.5} & \multicolumn{1}{c|}{88.5} & 89.0          \\
Task 3   & \multicolumn{1}{c|}{81.0} & \multicolumn{1}{c|}{86.5} & \multicolumn{1}{c|}{86.0} & \multicolumn{1}{c|}{87.0} & \multicolumn{1}{c|}{79.0} & \multicolumn{1}{c|}{\textbf{87.5}} & \multicolumn{1}{c|}{86.5} & \multicolumn{1}{c|}{86.5}          & \multicolumn{1}{c|}{79.0} & \multicolumn{1}{c|}{86.5} & \multicolumn{1}{c|}{85.5} & 87.0          \\
Task 4   & \multicolumn{1}{c|}{89.0} & \multicolumn{1}{c|}{90.5} & \multicolumn{1}{c|}{92.0} & \multicolumn{1}{c|}{91.5} & \multicolumn{1}{c|}{89.0} & \multicolumn{1}{c|}{91.0}          & \multicolumn{1}{c|}{91.5} & \multicolumn{1}{c|}{\textbf{92.5}} & \multicolumn{1}{c|}{89.5} & \multicolumn{1}{c|}{91.5} & \multicolumn{1}{c|}{91.5} & 91.5          \\
Task 5   & \multicolumn{1}{c|}{84.0} & \multicolumn{1}{c|}{89.0} & \multicolumn{1}{c|}{90.5} & \multicolumn{1}{c|}{90.5} & \multicolumn{1}{c|}{84.0} & \multicolumn{1}{c|}{89.0}          & \multicolumn{1}{c|}{91.0} & \multicolumn{1}{c|}{91.5}          & \multicolumn{1}{c|}{85.0} & \multicolumn{1}{c|}{87.5} & \multicolumn{1}{c|}{91.5} & 91.5         
\end{tabular}
\end{table*}

\subsection{Baseline}
\label{sssec:baseline}

\noindent For the baseline methods, 
two linear classifiers, logistic regression and support vector machines, and one non-linear classifier, random forest, are tested 
with different BOW representations
such as TF (term-frequency), TF-Nome (TF normalized by the document size), Binary (boolean occurrence value), and Tf-IDF.
Table~\ref{tbl:baseline} shows the accuracy measures for the five classification tasks with different combinations of classifiers and vector representations.
On average, SVM using TF-IDF outperforms the other baseline models.

\subsection{Convolutional Neural Networks Model}
\label{sssec:cnnmodel}

\noindent Since our work is the first to apply
a CNN model to document classification in radiology reports,
the goal of the experiments with CNN 
is to confirm the hypothesis that having 
big data in neural models is beneficial.
In combination with other factors, this motivation led us to vary the following three hyper-parameters.
Throughout the experiments, we set 
the number of filters to 64, 
the drop-out rate to 0.2.
We also used four kinds of filters with different sizes
 which are
$2\times d$, $3\times d$, $4\times d$ and $5\times d$ 
with various $d$:

\vspace{1ex}
\begin{enumerate}
\item The dimension of vectors of word2vec\\: 100, 200, and 400.
\item The number of documents used for embedding training\\: 20k, 40k, 60k, and 80k.
\item Optimization methods for embedding training\\: SKIP and CBOW.
\item The number documents for CNN training\\: 500 and 1,000.
\end{enumerate}
\vspace{1ex}

\begin{table*}[!ht]
\centering
\caption{Accuracy (in \%) of the NAM with different 
sets of hyperparameters measured on the development set.
AM-NUM represents the number of filters 
when creating an attention matrix described in Figure~\ref{fig:eav1}.
The best model for each task is marked as bold text.
}
\label{tbl:nam}
\begin{tabular}{c||c|c|c|c|c|c|c|c|c|c|c|c}
W2V-DIM  & \multicolumn{4}{c|}{100}                              & \multicolumn{4}{c|}{200}                                  & \multicolumn{4}{c}{400}                              \\ \hline
W2V-TASK & \multicolumn{2}{c|}{SKIP} & \multicolumn{2}{c|}{CBOW} & \multicolumn{2}{c|}{SKIP}     & \multicolumn{2}{c|}{CBOW} & \multicolumn{2}{c|}{SKIP} & \multicolumn{2}{c}{CBOW} \\ \hline
AM-NUMFIL   & 10     & 20             & 10         & 20         & 10           & 20           & 10     & 20             & 10         & 20         & 10         & 20         \\ \hline \hline
Task 1   & 88.0    & 88.5            & 88.5        & 89.0        & \textbf{90.0} & 89.5          & 89.5    & 88.0            & 87.5        & 88.0        & 88.0        & 88.5        \\
Task 2   & 87.5    & 88.0            & 88.5        & 88.5        & \textbf{89.0} & \textbf{89.0} & 88.0    & 88.5            & 88.5        & 88.5        & 88.0        & 88.0        \\
Task 3   & 86.0    & \textbf{88.0}   & 87.5        & 87.0        & 85.0          & 85.0          & 85.0    & 85.5            & 86.0        & 86.5        & 87.5        & 85.5        \\
Task 4   & 92.0    & 91.5            & 93.0        & 92.5        & 92.0          & 91.5          & 92.5    & \textbf{93.5}   & 92.0        & 91.5        & 92.5        & 92.0        \\
Task 5   & 92.0    & \textbf{93.0}   & 92.5        & 92.0        & 92.5          & 91.0          & 91.0    & 92.5            & 92.0        & 91.0        & 90.5        & 91.0       
\end{tabular}
\end{table*}

\begin{figure*}[htb]
\centering
\subfigure[In word2vec training, as the number of documents increases, 
the resulting vectors are more effective in training classifiers.]
{\label{fig:pc1}\includegraphics[width=0.3\linewidth]{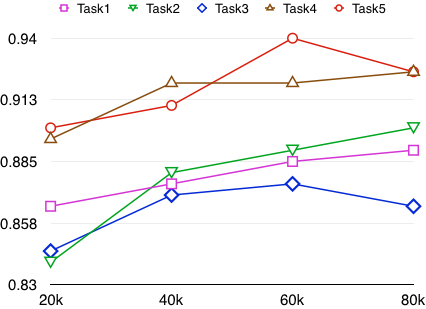}}
\subfigure[As the dimension of word embeddings increases, 
the performance marginally increases. The dimension of 100 always produces lower accuracies.]
{\label{fig:pc2}\includegraphics[width=0.3\linewidth]{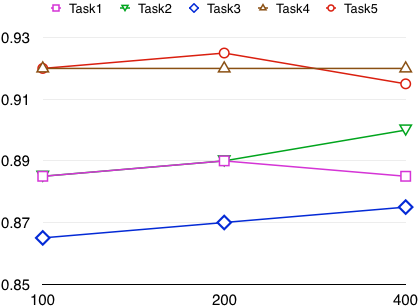}}
\subfigure[Large set of training documents is definitely effective for learning.]
{\label{fig:pc3}\includegraphics[width=0.3\linewidth]{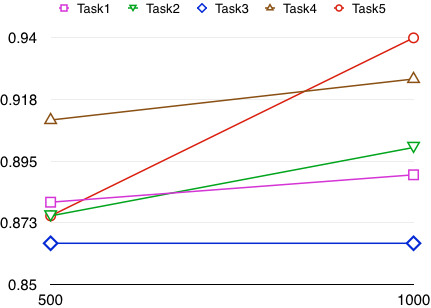}}
\caption{Performance changes of the CNN model across various sets of hyperparameters.}
\label{fig:pc}
\end{figure*}

\noindent Table~\ref{tbl:cnn} shows all experimental results except for the last parameter indicating the size of the CNN training data.
We exclude the effect of the last hyperparameters because the models trained with the larger number of data
always perform better than the ones trained with smaller number  of data
as shown in Figure~\ref{fig:pc3}.
According to Table~\ref{tbl:cnn},
besides the effect of the number of annotated documents,
three performance tendencies of the CNN models are conspicuous.
The first finding is that
a larger word2vec dimension is advantageous in performance,
as shown in Figure~\ref{fig:pc2}.
All the best models are incorporated with
the word2vec dimension of 
either 200 or 400 (note that no best model is integrated with the dimension of 100).
The reason of this finding is that 
projecting to a smaller dimensional space usually 
requires loss of information.

Secondly,
abundant (unannotated) documents for word2vec training increase the accuracy, as presented in Figure~\ref{fig:pc1}.
Since the purpose of word2vec is to find 
proper word representations based on the context words,
a rich source of training data is helpful
to find precise projections.
Another general trend is that
word2vec with the SKIP method produces more accurate results
than the one with the CBOW method.
As Mikolov
stated 
in the discussion group
\footnote{T. Mikolov, ``Differences between the skip-gram
and the cbow models'' in a google group discussion.}, 
SKIP method generally works better than CBOW
if the training data is small.
The radiology report dataset can be considered as 
small dataset compared to a large general corpus,
such as Wikipedia that consists of text in the millions.

The best models for each task
are selected based on the maximum scores 
evaluated on the development set,
which are marked as bold faced numbers in
Table~\ref{tbl:cnn}.
To compare performances with the baseline,
the five selected models are evaluated on the test set.
The test scores for the five tasks are
88.0, 86.5, 85.0, 89.5, and 87.0, 
all of which are included in the model comparison table 
(Table~\ref{tbl:allcomparison}).

\subsection{Neural Attention Model}
\label{sssec:neural-attention-model}

\setcounter{subfigure}{0}
\begin{figure*}[htb]
\centering
\subfigure[Heatmap of an radiology report for the task 2
whose purpose is to classify patients with acute intracranial bleed.
Words that describe or imply bleeding get higher attention than other less important words. For example, 
the NAM mostly focused on the words "intraparenchymal hemorrhage"
to classify if the radiologist has noticed an acute bleed and described it in the radiology report.
]
{\label{fig:hmap1}\includegraphics[width=\linewidth]{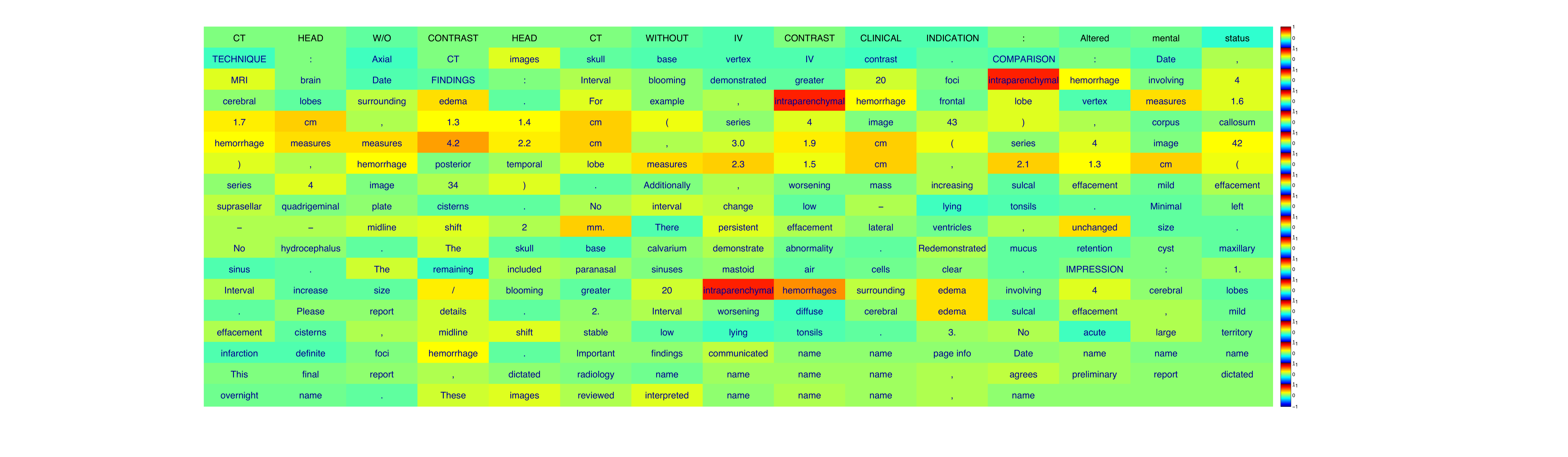}}
\subfigure[Heatmap of an radiology report for task 3
whose purpose is to classify patients with a mass effect.
Mass effects denotes swelling of one or more parts of the brain and results in compression of other regions inside the cranium, such as the remainder of the brain, blood vessels, and vital cranial nerves.
Radiologists describe mass effect in many ways of which "sulcal effacement" is a major description; the NAM puts significant attention on this term. 
]
{\label{fig:hmap2}\includegraphics[width=\linewidth]{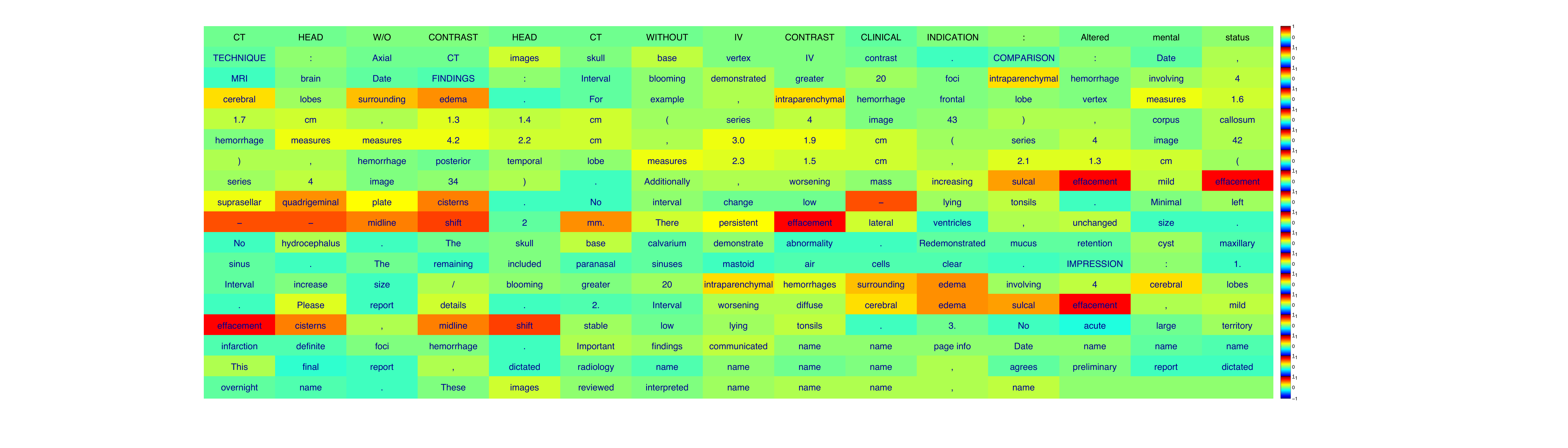}}
\caption{Comparison of heatmaps for two tasks. 
Important Keywords for 
the corresponding purpose of each task
draw more attention. All personal information such as date and names (of a patient and a doctor) are deidentified as 'date' and 'name', respectively. }
\label{fig:hmap}
\end{figure*}

\noindent The motivation for applying an attention mechanism 
to the CNN model is to retrieve rationales
of prediction results.
In order to extract this information from
learned model,
the analysis of the embedding attention vector (EAV) in Figure~\ref{fig:eav1}
should be performed.
Since the EAV conveys weights of each token in a document,
it can be considered as the concentration factor
which reflects the degree of attention of the machine
when it performs a classification task.
We visualize two heat maps to clarify 
what words the machine focused on depending on the tasks
in Figure~\ref{fig:hmap}.
If a task is to classify 
patients with bleeding,
the machine should focus on bleeding indicative words,
such as ``intraparenchymal hemorrhage", as shown in Figure~\ref{fig:hmap1}. 

In contrast, in Figure~\ref{fig:hmap2}, 
if the machine performs
a task of classifying patients with mass effect,
it should focus on different key words, such as 
``sulcal effacement", although the text is the same.
To compare performances with the baseline,
we select five attention models
that perform the best for each task 
when evaluated on the development set.
This result is summarized in Table~\ref{tbl:nam}.
These five selected models are evaluated on the test set
to compare with other models.
The scores are
88.0, 87.5, 85.0, 87.5, and 87.0, in order of the tasks,
all of which are included in Figure~\ref{tbl:allcomparison}.

\subsection{Performance Comparison}
\label{sssec:performance-comparison}

\noindent As shown in Table~\ref{tbl:allcomparison},
the proposed models outperform the baseline.
Both of the neural models 
gained more than 3\% improvements on average.
We can estimate the superiority of our models
by comparing the accuracies with the agreement scores 
between the two human annotators.
As noted in Section~\ref{ssec:corpora},
two annotators labeled the documents 
according to each task.
Since there are discrepancies between two experts,
we measured the agreement scores.
Although these scores are not directly comparable to 
the accuracies,
we can assess the proposed model based on theses scores.

\begin{table}[!ht]
\centering
\caption{Accuracy comparison (in \%) on the test data. 
The two proposed models outperform the baseline.
Furthermore, they achieve higher accuracies than 
human agreement scores in three tasks}
\label{tbl:allcomparison}
\begin{tabular}{c||c||c|c|c}
\multirow{2}{*}{} & \multirow{2}{*}{Human Agreement} & \multicolumn{3}{c}{Accuracy}                        \\ \cline{3-5} 
                  &                                  & SVM (Baseline) & CNN              & NAM              \\ \hline \hline
Task 1            & 86.5                            & 85.5          & \textbf{88.0}  & \textbf{88.0}  \\
Task 2            & 86.5                            & 83.0           & 86.5            & \textbf{87.5} \\
Task 3            & 81.5                            & 83.5          & \textbf{85.0}  & \textbf{85.0}  \\
Task 4            & 94.0                             & 85.5          & \textbf{89.5} & 87.5            \\
Task 5            & 90.0                              & 81.0           & \textbf{87.0}  & \textbf{87.0}  \\ 
\end{tabular}
\end{table}

\noindent In task 1, task 2, and task 3, 
our models achieved higher accuracies than human agreement scores.
If we compare between the two proposed models,
although the performance of the two proposed models are 
approximately identical,
NAM is more desirable because of its useful byproduct 
(attention information).

\section{Conclusion}

\noindent This paper proposes two neural models 
that effectively apply CNN and attention mechanism 
to a medical document classification problem, namely radiology reports.
Our experiments show that 
the proposed models can 
not only improve accuracy compared to non-neural models,
but also enable interpretability to a neural model.
The experiments on various combinations of hyperparameter 
show that neural models are effective on large dataset.
The attention heatmap analysis confirms that 
the attention mechanism endows CNN models with explanatory features, 
which gives good rationales of the given prediction.

The proposed attention models are applied to each single word. However, focusing on multiple words could give more promising information.
Application of the attention mechanism to multiple words at the same time is a possible direction.
Since we focused on a simple and yet well performing system,
ensemble of multi-layer CNN models could be applied
in order to maximize the score.


\section*{Acknowledgment}

\noindent We gratefully acknowledge the support of the Foundation of the American Society of Neuroradiology (ASNR) Comparative Effectiveness Research (CER) Grant, the Association of University Radiologists (AUR) General Electric Radiology Research Academic Fellowship (GERRAF) Grant, and the Infosys Research Enhancement Grant.
A special thank is due to Jung-Hyun Kang for assisting to generate the figures.



\bibliographystyle{IEEEtran}
\bibliography{radatt}
%



\end{document}